\pdfoutput=1
\documentclass[11pt]{article}

\usepackage[preprint]{acl}

\usepackage{times}
\usepackage[T1]{fontenc}
\usepackage[utf8]{inputenc}

\usepackage{microtype}

\usepackage{inconsolata}

\usepackage{graphicx}

\usepackage{hyperref}
\usepackage{url}
\usepackage{soul}
\usepackage{pifont}
\usepackage{xcolor}
\usepackage{colortbl}
\usepackage[most]{tcolorbox}
\usepackage{pifont}
\usepackage{algpseudocode}
\usepackage{amsfonts}
\usepackage{fdsymbol}
\usepackage{graphicx}
\usepackage{subfig}
\usepackage{mathtools}
\usepackage{amsmath}
\usepackage{amsthm}
\usepackage{listings}
\usepackage{wrapfig}
\definecolor{light_gray}{rgb}{.95,.95,.95}
\definecolor{custompurple}{RGB}{93,0,93}
\definecolor{customorange}{RGB}{255,132,6}
\definecolor{customgold}{RGB}{213,177,52}
\definecolor{customblue2}{RGB}{28,205,188}
\definecolor{no_persona_color}{RGB}{152,226,245}
\definecolor{persona_color}{RGB}{193,167,246}
\usepackage{booktabs}
\usepackage{adjustbox}
\usepackage{scalerel}
\usepackage{fontawesome5}
\usepackage{booktabs,tabularx,array,colortbl,xcolor}
\usepackage{siunitx}
\usepackage[most]{tcolorbox}
\usepackage{url}
\usepackage{twemojis}
\usepackage{dsfont}
\usepackage{bm}
\usepackage{bbm}
\usepackage{graphicx}
\usepackage{color}
\usepackage{multicol}
\usepackage{multirow}
\usepackage{wrapfig,lipsum,booktabs}
\usepackage[ruled,vlined,linesnumbered]{algorithm2e}
\usepackage{pgfplots}
\pgfplotsset{compat=1.12}
\usepackage{xcolor}
\usepackage{tikz}
\usepackage{xspace}
\usepackage{makecell}
\usepackage{soul}
\usepackage{amsmath,amsfonts}
\usepackage{subcaption}
\usetikzlibrary{calc}
\usepgfplotslibrary{groupplots}
\usetikzlibrary{angles,quotes} 
\usetikzlibrary{shapes,arrows}
\usetikzlibrary{backgrounds}
\usetikzlibrary{matrix}
\usepackage{tikz-3dplot}
\usepackage{hyperref}
\usepackage{cleveref}
\usepackage{paralist}
\usepackage{cancel}
\usepackage{xspace}
\usepackage{todonotes}
\usepackage{tabu}
\usepackage{rotating}
\usepackage{etoolbox}
\usepackage{adjustbox}
\usepackage{enumerate}
\usepackage{enumitem}
\setitemize{noitemsep,topsep=0pt,parsep=0pt,partopsep=0pt}
\setenumerate{noitemsep,topsep=0pt,parsep=0pt,partopsep=0pt}
\usepackage{pifont}
\usepackage{cancel}
\usepackage{lipsum}
\usepackage{listings,lstautogobble}
\usepackage{fancyvrb}
\usepackage{fvextra}
\usepackage{caption}
\usepackage{pgf-pie} 
\usepackage{array, makecell}
\usepackage{wrapfig}

\definecolor{codegreen}{rgb}{0,0.6,0}
\definecolor{codegray}{rgb}{0.5,0.5,0.5}
\definecolor{codepurple}{rgb}{0.58,0,0.82}
\definecolor{backcolour}{rgb}{0.95,0.95,0.92}

\definecolor{citysimlight}{RGB}{238,245,255}   % Light, airy blue background
\definecolor{citysimaccent}{RGB}{88,140,255}
\definecolor{CrowdCol}{HTML}{5A6EFF}   % vivid blue
\definecolor{ExpertCol}{HTML}{50F05A}  % bright green
\definecolor{GPTCol}{HTML}{FF2828}     % strong red
\definecolor{ClaudeCol}{HTML}{FA8C32}  % pumpkin orange

\definecolor{chart Idle}{gray}{.6}
\definecolor{chart Poor}{RGB}{242,28,28}
\definecolor{chart Ok}{RGB}{248,172,37}
\definecolor{chart Ideal}{RGB}{1,151,0}
\definecolor{chart Over}{RGB}{0,125,234}

\newdimen\tempdim

\newcommand*{\ChartLegend}[1]{%
  \ifdim\lastkern=1sp %
    \hspace{1em}%
  \fi
  \ChartBox{0.75em}{#1}%
  \,#1%
  \kern-1sp\kern1sp\ignorespaces
}
\newcommand*{\ChartBox}[3]{%
  \begingroup
    \settoheight{\tempdim}{L}%
    \edef\tempheight{\the\tempdim}%
    \settodepth{\tempdim}{g}%
    \edef\tempdepth{\the\tempdim}%
    \tikz[
      baseline=0pt,
      inner sep=0pt,
    ]
    \node[
      fill={#3!#2},
      rounded corners=1pt,
      anchor=base,
    ]{%
      \vphantom{g\"A}%
      \pgfmathsetlength{\tempdim}{#1}%
      \kern\tempdim\relax
    };%
  \endgroup
}

\title{CitySim: Modeling Urban Behaviors and City Dynamics with Large-Scale LLM-Driven Agent Simulation}

\author{
 \textbf{Nicolas Bougie\textsuperscript{1}},
 \textbf{Narimasa Watanabe\textsuperscript{1}}
 \\ \texttt{\{nicolas.bougie,narimasa.watanabe\}@woven.toyota}\\
\\
 \textsuperscript{1}Woven by Toyota
}

\begin{document}
\renewcommand*{\arraystretch}{1.2}
\newcommand*{\chart}[3]{%
  \ChartBox{20mm/3000*(#1-400)}{#2}{#3}%
}
\newcommand*{\humanchart}[3]{%
  \ChartBox{20mm/1300*(#1-700)}{#2}{#3}%
}
\maketitle
\begin{abstract}
Modeling human behavior in urban environments is fundamental for social science, behavioral studies, and urban planning. Prior work often rely on rigid, hand-crafted rules, limiting their ability to simulate nuanced intentions, plans, and adaptive behaviors. Addressing these challenges, we envision an urban simulator (CitySim), capitalizing on breakthroughs in human-level intelligence exhibited by large language models. In CitySim, agents generate realistic daily schedules using a recursive value-driven approach that balances mandatory activities, personal habits, and situational factors. To enable long-term, lifelike simulations, we endow agents with beliefs, long-term goals, and spatial memory for navigation. CitySim exhibits closer alignment with real humans than prior work, both at micro and macro levels. Additionally, we conduct insightful experiments by modeling tens of thousands of agents and evaluating their collective behaviors under various real-world scenarios, including estimating crowd density, predicting place popularity, and assessing well-being. Our results highlight CitySim as a scalable, flexible testbed for understanding and forecasting urban phenomena.
\end{abstract}

\section{Introduction}
Simulating realistic city-scale behaviors is a longstanding goal in social science and artificial intelligence ~\cite{hofman2021integrating,lazer2009computational}. Traditional models often leverage hand-crafted rules or utility functions, restricting their ability to represent the diversity, adaptability, and long-term dynamics inherent in humans ~\cite{zheng2022ai,emnlp/WangCC23}. As a result, they often yield unrealistic behaviors: rigid schedules, inability to develop new preferences, and poor adaptation to novel or changing environments ~\cite{feng2024citybench}. Hence, it remains challenging to simulate nuanced psychological, social, and environmental drivers shaping actions in urban settings.

Recent advancements in large language models (LLMs) offer promising avenues in creating human-like agents for studying complex urban phenomena, enabling detailed exploration into individual-level decisions and social interactions ~\cite{epstein1999agent,macal2005tutorial,gao2024large,park2023generative}. Numerous studies have pointed out that after being empowered by LLMs, agents gain the ability to reason, plan, and interact through natural language ~\cite{gao2024large,wei2022chain,bougie2025simuser,li2023camel,park2023generative, bougie2024generative}. A pioneering example is the Generative Agent \cite{park2023generative}, which constructs a small-scale society within a 2D game engine. Recently, AgentSociety \cite{piao2025agentsociety} has introduced a large-scale multi-agent platform, but lacks long-term cognitive mechanisms such as evolving goals, evolving preferences, or spatial memory. Despite advances in modeling motivation, memory, and social interaction~\cite{ye2025mobilecity}, current approaches still face several challenges. (1) Agents often plan activities in a fixed sequential manner, overlooking the relative importance and interdependencies among daily tasks; (2) Long-term simulation is underexplored, ignoring the effect of beliefs and long-term goal formation and adaptation. (3) Place selection and vehicle choice are usually not considered.

In this work, we present \textbf{CitySim}, a scalable city simulation framework empowered by LLMs. CitySim agents autonomously generate daily schedules and long-term plans through a recursive, value-driven planning process that balances mandatory activities, personal habits, and situational context. Each agent is equipped with spatial and temporal memories, which enable agents to recall past experiences, form and update beliefs about places they visit, and adapt their future decisions accordingly. Namely, agents continuously revise beliefs, transportation choices, and internal states (such as mood or satisfaction) in response to needs, environmental feedback, and social interactions. To lay a reliable and diverse foundation, agents are equipped with a persona module derived from real-world surveys, encompassing demographic, personality traits, preferences and habits. This results in heterogeneous, context-aware urban populations capable of capturing the complexity of real societies.

\section{Related Work}

Replicating human behavior in urban environments remains a challenge ~\cite{hofman2021integrating,lazer2009computational}. Traditional agent-based models have been widely used to study complex phenomena, resource allocation, and policy evaluation~\cite{epstein1999agent,macal2005tutorial, wilensky2015introduction}. Yet, they depend on hard-coded rules or fixed utility functions, constraining their capacity to capture behavioral diversity, adaptability, and long-term dynamics~\cite{feng2024citybench,zheng2022ai,emnlp/WangCC23}. Recent frameworks such as CityBench~\cite{feng2024citybench} and AI4SIM~\cite{zheng2022ai} have begun to integrate richer, data-driven approaches, yet realistic cognitive and motivational modeling remains limited.

Recently, LLMs have opened new possibilities for simulating human-like agents in virtual worlds~\cite{park2023generative,gao2024large,li2023camel,wei2022chain}. LLM-powered agents can reason, plan, and interact through natural language~\cite{gao2024large,wei2022chain,bougie2025simuser,park2023generative, bougie2024generative}. Nevertheless, most LLM-based agents still plan activities in a myopic or rigid manner, struggle to update beliefs from experience, and often represent personas solely via demographics~\cite{emnlp/WangCC23}. To support different application scenarios, \citet{corr/abs-2309-07870} present an open-source framework for autonomous language agents, and \citet{iclr/HongZCZCWZWYLZR24} demonstrate how agents can collaborate in complex software engineering tasks. As research moves toward larger-scale simulations, computational efficiency becomes crucial. \citet{corr/abs-2411-10109} scale up simulations to 1,000 agents but still inherits prohibitive costs. Frameworks such as AgentSociety~\cite{piao2025agentsociety} or MobileCity \cite{ye2025mobilecity} use episodic memory and gravity-based place selection, but neglect habits and transport choices. CitySim advances this area by integrating recursive activity planning, dynamic memory and belief modules, enabling lifelike, context-sensitive, long-term behaviors.

\section{Method}
We introduce \textbf{CitySim}, a framework simulating human-like behavior in a dynamic, graph-structured urban environment. Agents are endowed with advanced cognitive representations, including personas, long-term goals, beliefs, and needs. At each simulation step, an agent first \textbf{perceives} its environment (e.g., location, presence of friends) and internal state. Along with insights retrieved from a memory module, it decides on a course of action -- seeking food when hungry. The agent then \textbf{reacts} by either following its current plan, adjusting future activities when an unexpected event happens, or filling the agent's free time via value-driven planning. Finally, it \textbf{reflects} on recent experiences to develop beliefs, opinions and habits, synthesizing memories into higher-level reflections.

\subsection{Cognitive State Representation}

\subsubsection{Persona Module}
\label{sec:persona_module}
The persona module is fundamental for aligning agents with genuine human behaviors in urban simulation. To lay a reliable foundation, questionnaire-derived attributes include:
\begin{itemize}
    \item \textbf{Demographic Attributes:} Name, age, gender, occupation, income, hobbies, education, household composition, and life stage. These modulate the agent’s activity space (e.g., children attend school; retirees prefer daytime leisure) and shape patterns in daily routines.
    \item \textbf{Spatial Anchors:} Home, and work/school locations.
    \item \textbf{Psychographic Traits:} Personality traits are defined by the Big Five personality facets: \textit{Openness}, \textit{Conscientiousness}, \textit{Extraversion}, \textit{Agreeableness}, and \textit{Neuroticism}, each measured on a 1 to 3 scale.
    \item \textbf{Habits and Preferences:} routines that underlie human activity. Each agent is initialized with a set of empirically-derived habits and preferences, including activity preferences, habits (e.g., early riser), and leisure patterns. 
\end{itemize}

\subsubsection{Memory Module}
\label{sec:spatial_memory}
Humans retain diverse memories, bifurcating mainly into factual and emotional categories \cite{labar2006cognitive}. In CitySim, the memory module comprises three components: \textbf{temporal}, \textbf{reflective}, and \textbf{spatial} memories.

\noindent\textbf{Temporal Memory} This memory is organized chronologically, with multiple memory nodes in each stream. Each memory node contains four components: \textbf{time}, \textbf{location}, and \textbf{observation}, and \textbf{key}. The key serves to filter observations during retrieval. New entries are appended at each simulation step, action, or reflection.

\noindent\textbf{Reflective Memory} The reflective memory records the agent’s thoughts and attitudes toward events stored in temporal memory. Each entry is linked to one or more nodes in the temporal memory, reflecting how the agent perceives or reacts to a specific event. At the end of each day, we synthesize those memories into higher-level reflections, enabling the agent to draw conclusions about itself.

\noindent\textbf{Spatial Memory.} Spatial memory maintains beliefs $b_i \in \mathbb{R}^K$ about point of interests (POIs), where each dimension corresponds to an aspect $\in\{\textit{price}, \textit{atmosphere}, \textit{satisfaction}, \textit{convenience}\}$. When the agent visits a POI $i$ at time $t$ and observes outcome $o_i^{(t)}$, the belief $b_i^{(t)}$ is updated using a Kalman filter to reduce noise (see Appendix). If a POI $i$ has not been visited, $b_i$ is initialized via embedding-based similarity to previously visited locations: $b_{i} \leftarrow \mathbb{E}_{j \in \mathcal{N}(i)}[b_{j}]$, where $\mathcal{N}(i)$ denotes similar POIs retrieved from spatial memory. Note that the uncertainty is retrieved along $b_{i}$ to guide the agent during the POI selection. To reflect forgetting and environmental changes, we introduce a \textbf{decay}. After a simulated day, each dimension $d$ of $b_i$ is updated as: $b_{i,d} \leftarrow (1 - \lambda) b_{i,d} + \lambda b_{0,d}$, where $b_0$ is the neutral value ($0.5$) and $\lambda$ is the decay rate.

\subsubsection{Belief Module}
\label{sec:belief}
This module is triggered each time the agent visits a place. Upon visiting POI $i$, the agent generates a subjective observation $o_i^{(t)}$ by prompting an LLM with visit-specific context, including the agent’s persona, current activity, emotional state, and description of the POI. This observation captures the agent’s immediate appraisal of the visit, providing a multi-dimensional assessment with associated reasoning. The new observation is then integrated with the prior belief $b_i^{(t-1)}$, as described above.

\subsubsection{Needs Module}
\label{sec:needsblock}
The agent’s tracks and prioritizes four primary needs: hunger, energy, safety, and social connection. At the \textbf{start} of each day, an LLM prompt, conditioned on demographic and temporal context, serves to initialize scores: $\mathbf{s}_0 = \{\text{hunger},\, \text{energy},\, \text{safety},\, \text{social}\} \in [0, 1]^4$. Throughout the day, scores \textbf{decay} continuously based on decay rate $\alpha_n$ for need $n$: $s_n(t) = \max\left(0,\, s_n(t-\Delta t) - \alpha_n \Delta t\right)$. Following each activity or significant events, the LLM evaluates outcomes and updates scores based on agent experience, contextual information, and current internal state. Needs are prioritized (\texttt{hungry} $>$ \texttt{safe} $>$ \texttt{tired} $>$ \texttt{social}) using thresholds $T_n$: $\textit{dominant need} = \arg\min_n \left\{ s_n \leq T_n \right\}$. If a higher-priority need arises, ongoing plans may be interrupted and needs reprioritized. The dominant need is stored as plain text in the persona module.

\subsubsection{Long-Term Goal Module}
\label{sec:longtermgoal}
We model the formation and revision of the agents’ high-level aspirations, drawing on the Maslow’s Hierarchy of Needs \cite{huitt2007maslow}. Goals are revisited monthly or following major life events (e.g., employment changes). In detail, the LLM is queried using persona, financial status, social contacts, recent activities, and current goals. Along with these inputs, we compute the \textbf{need fulfillment}, defined as the proportion of the day during the needs exceed their thresholds. We also monitor \textbf{financial stress} ($\texttt{income} < 0.9 \times \texttt{expenses}$) and \textbf{social isolation} (fewer than 3 unique contacts in last 7 days). To capture the \textbf{interest}, we calculate the proportion of recently visited POIs whose current satisfaction belief exceeds 0.5: $\text{interest} = \frac{1}{|\mathcal{V}|} \sum_{i \in \mathcal{V}}{} \mathbb{I}[b_{i}^{\text{sat}} > 0.5]$, where $b_i^{\text{sat}}$ is the satisfaction belief for POI $i$ and $\mathcal{V}$ is the set of POIs visited in the last 30 days. Given this context $c$, a structured prompt $p_{\text{goal}}$ conditions the LLM to generate coherent short (few weeks) and long-term goals: $g_t^1, g_t^2, \dots, g_t^M \sim \texttt{LLM}(p_{\text{goal}} \mid \theta, c)$, where resulting goals $g_t^m$ inform subsequent planning modules.

\subsubsection{Perception Module}
At each simulation timestep, the perception module receives an observation from the environment and determines whether the agent should react. If so, the perception module enumerates the set of available modules, each accompanied by a functional description, and queries the LLM to select the most appropriate module for the current situation. Module selection is managed by a dispatcher, which invokes the corresponding module (e.g., planning, social interaction) based on the agent’s inferred needs, short explanation, and required parameters.

\subsection{Mobility Behaviors}

\subsubsection{Planning Module}
\label{sec:planning}
Daily schedules are generated via a \textbf{recursive decomposition} of time into \texttt{[blocks]}, each including a starting time, duration, and activity/intention. Planning begins each day with mandatory tasks (e.g., sleep, work) based on each agent persona and needs, then recursively fills remaining \texttt{[EMPTY]} blocks with medium-priority tasks (e.g., meals, hygiene). If a selected activity does not fill the entire interval, the block may be subdivided according to the activity duration.

Some \texttt{[blocks]} remain unfilled after this process. According to Maslow’s Hierarchy~\cite{huitt2007maslow}, these are typically used for leisure or long-term goals (e.g., hobbies, socializing), and are filled at execution time based on the agent’s state, location, needs, and schedule using \textbf{value-driven planning}. For each empty block, the agent generates and evaluates multiple candidate activities, selecting the one expected to best satisfy intrinsic desires. This selection is handled through a single structured LLM call with internal reasoning steps.

\subsubsection{Place Selection Module}
\label{sec:placeselection}
For each activity, the location is determined using a belief-aware gravity model, extending AgentSociety~\cite{piao2025agentsociety}. For home or work activities, addresses from the agent’s persona are used.

\noindent\textbf{Step 1: Macro-level Area Selection.} The agent decides whether to remain in the vicinity or travel farther by prompting the LLM with intention, schedule, emotional state, area visit history, and popular nearby areas (ranked by distance and popularity).

\noindent\textbf{Step 2: Micro-level POI Selection.} Within the chosen area, the agent:
\\\textbf{Intention Extraction:} Determines required POI types (e.g., café, park) and adjust feasible ranges by integrating internal (e.g., age, daily schedule) and environmental factors (e.g., weather, traffic), providing a set of POIs candidates.
\\\textbf{Belief-weighted Gravity Model:} For each candidate POI $i$, computes the selection weight as
\begin{equation}
    p_{ij} = \frac{(b_j +  \varepsilon)\,/\, D_{ij}^{\,1+\gamma(b_j - 0.5)}}
                  {\sum_k (b_k+ \varepsilon) \,/\, D_{ik}^{\,1+\gamma(b_k - 0.5)}}
\end{equation}
where \(b_j\) is the belief-based attractiveness of location \(j\), and \(D_{ij}\) is the distance, $\gamma$ controls distance decay, and \( \varepsilon \) is a small positive constant to ensure numerical stability. Here, \(b_j\) is the weighted sum of current beliefs. If no belief exists for \(j\), it is estimated from similar POIs in spatial memory.

\subsubsection{Vehicle Selection Module}
\label{sec:vehicleblock}
Finally, the most appropriate transport mode is estimated for each trip. Given trip context, including distance $d$ to the next POI, time of day $t$, month $m$, weather $w$, temperature $T$, and persona $p$, a structured prompt $p_{v}$ is used to query the LLM: $\texttt{LLM}(p_{v}~|~d, t, m, w, T, p, \theta, \mathcal{V})$, where $\mathcal{V}$ is the set of available vehicles. Along with the selected vehicle $v^*$, the agent is instructed to provide a brief justification, which are stored in the reflective memory. The process approximates $v^{*} = \arg\max_{v \in \mathcal{V}} U(d, t, m, w, T, p, \theta, \mathcal{V})$, where $U(\cdot)$ is an implicit utility function implemented by the LLM’s reasoning over the provided context.

\subsection{Social Behaviors}
\label{sec:social_module}
The foundation of our social module is a weighted \textbf{social network} where each edge encodes an agent's evolving social beliefs about others. Each agent $u$ maintains a social belief vector $b_{u,v}$ for every contact $v$, capturing dimensions: $\in \{\textit{affinity}, \textit{trust}, \textit{familiarity}\}$. These beliefs are initialized at simulation start based on demographic similarity and relationships, then updated continuously. We consider two types of interactions: \textbf{face-to-face} and \textbf{online}. After an interaction, $b_{u,v}$ is updated using observed outcome (positive, neutral, negative) for each dimension. 

\textbf{Face-to-face interactions} occur when agents are co-located in same space. Agent $u$ selects a conversation partner $v$ according to their current belief score, with probability: $p_v = \frac{b_{u,v}}{\sum_{v' \in V} b_{u,v'}}$, where $V$ is the set of eligible co-located agents, and $b_{u,v}$ is the current belief toward $v$. 

\textbf{Online interactions} simulate remote communication (e.g., phone calls or messasing). When agent $u$'s social satisfaction score falls below a defined threshold, it seeks to contact an acquaintance during leisure time, with selection probability based on beliefs and relationship strengths.

\section{Experiments}
\textbf{Settings.} Experiments are conducted using the urban simulation framework proposed in AgentSociety~\cite{piao2025agentsociety}. All agents are powered by the GPT-4o-mini version of ChatGPT, except when specified differently, with the number of agents set to 1,000 located in Tokyo metropolitan area.
\\\textbf{Baselines} We compare CitySim with GeAn \cite{uist/ParkOCMLB23}, AGA \cite{corr/abs-2402-02053}, HumanoidAgent \cite{emnlp/WangCC23}. We also report results with our closest competitors, MobileCity \cite{ye2025mobilecity} and AgentSociety \cite{piao2025agentsociety}.

\subsection{Macro-level Time Use}
\label{sec:exp_timeuse}

\begin{figure}[tbp]
    \centering
    \includegraphics[width=1.0\linewidth]{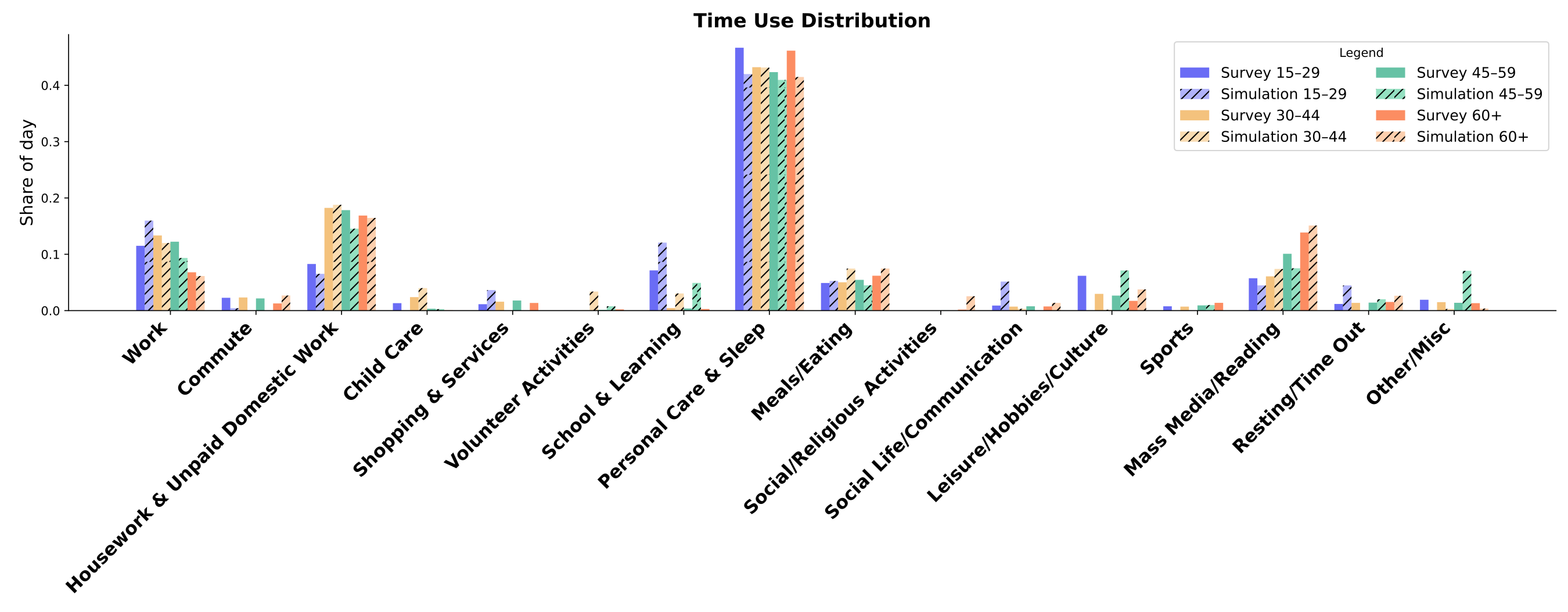}
    \caption{Time-use distribution across activity categories and age groups. Solid bars represent ground truth; striped bars show results from our simulation.}
    \label{fig:timeuse}
\end{figure}
To assess macro-level behavioral realism, we compare our agents’ time-use distribution with ground-truth data from the 2021 Japanese national time use survey~\cite{e-stat2021timeuse}. Each agent simulates two months of daily activities, which are mapped to the high-level activity categories used in the survey (e.g., Work, Commute, Housework, Personal Care \& Sleep). We aggregate and normalize the total time spent on each activity by age group, reporting the \emph{share of day}. As shown in Figure \ref{fig:timeuse}, the distribution of daily activities closely matches the survey statistics. These findings demonstrate that our model can faithfully produce city-scale, macro-level activity patterns.

\subsection{Pairwise Human Preferences}
\label{sec:parwise_likeliness}
\begin{figure}[tbp]
    \centering
    \includegraphics[width=1.0\linewidth]{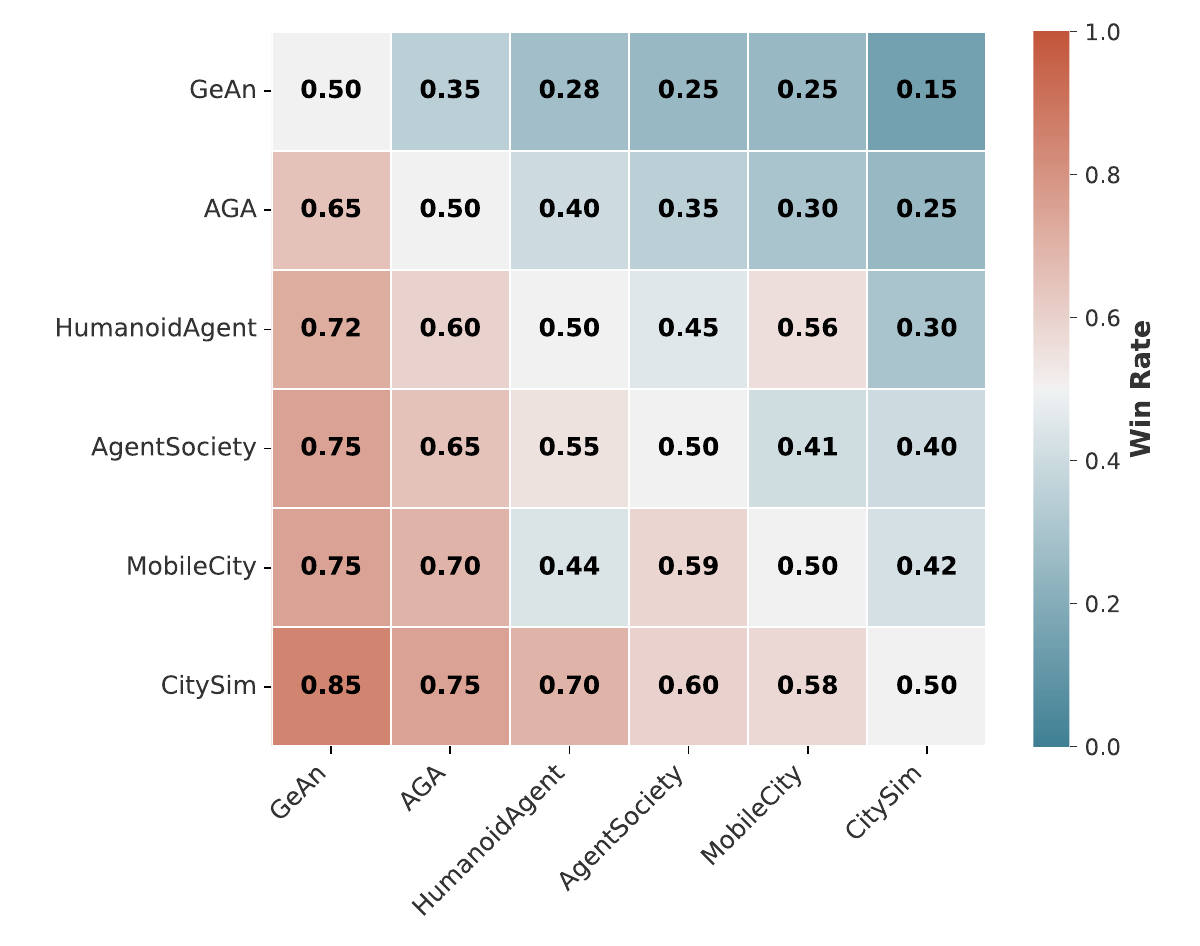}
    \caption{Pairwise win rate matrix. Each entry denotes the proportion of trials in which the row agent is judged more human-like than the column agent.}
    \label{fig:winrate}
\end{figure}

To evaluate the behavioral realism, we conducted 15 independent trials for each approach. To mitigate stylistic bias in generated activity sequences, all outputs were first normalized using Llama-3.1 70B to ensure consistent formatting across models. For each agent pair, GPT-4o was prompted to compare anonymized daily routines using three criteria: (i) \textbf{Naturalness}, (ii) \textbf{Coherence}; and (iii) \textbf{Plausibility}. The pairwise win rate (Figure~\ref{fig:winrate}) reflects how often each agent was judged more human-like than another. CitySim achieves the highest average win rates, outperforming all baselines. This is mainly due to explicit need modeling, dynamic goals, and memory-based planning, which support adaptive, context-sensitive behavior. In contrast, MobileCity and AgentSociety are more rigid and repetitive, often disregarding common social and temporal norms, contributing to suspicions of AI involvement.

\subsection{Travel Patterns}
\begin{figure}[tbp]
    \centering
    \includegraphics[width=1.0\linewidth]{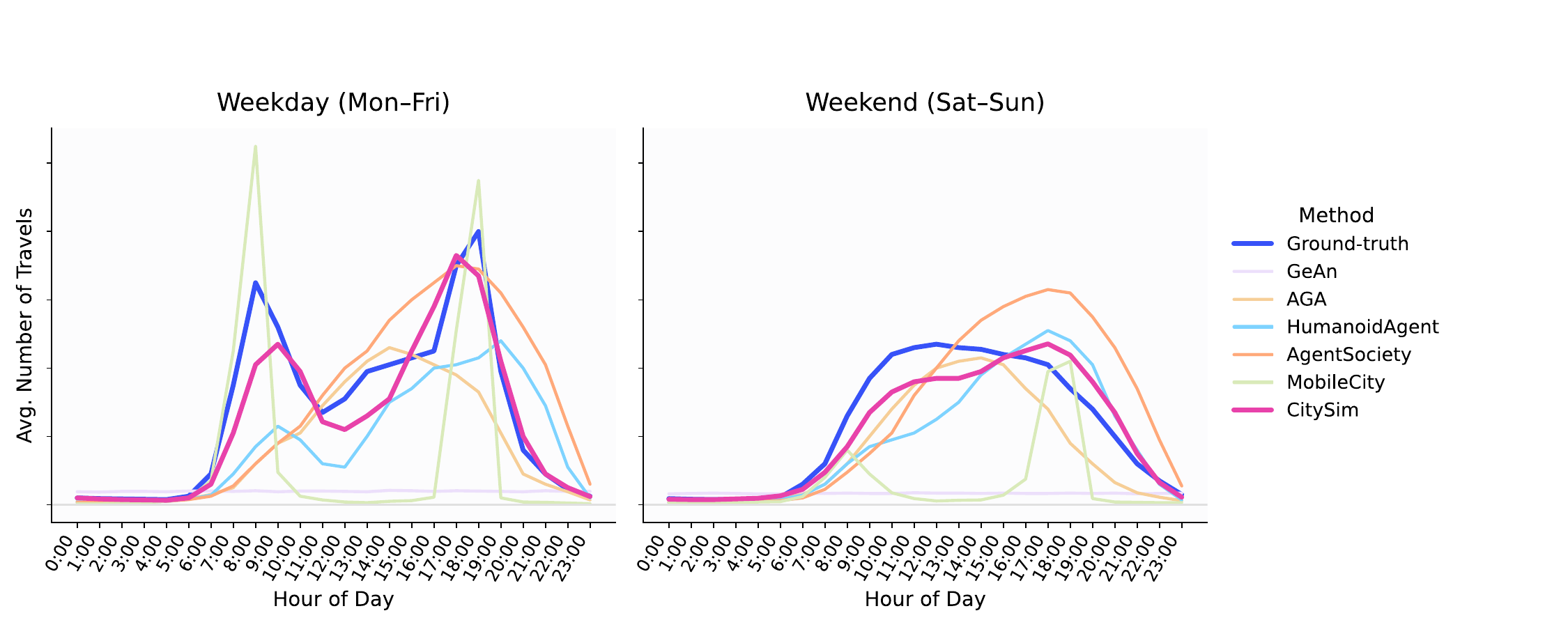}
    \caption{Average number of agent travels per hour on weekdays (left) and weekends (right).}
    \label{fig:mobility}
\end{figure}

We now compare simulated travel distributions with ground-truth data, derived from a proprietary city-scale dataset. Figure~\ref{fig:mobility} shows the average number of travels per hour for both weekdays and weekends. CitySim closely reproduces real-world patterns, matching the timing and amplitude of commuting peaks and weekend leisure activity. In contrast, MobileCity exhibits overly rigid spikes at commute times, while Baseline remains largely static. Other agent methods (AGA, HumanoidAgent, AgentSociety) capture some broad trends but display either diffused or mistimed travel peaks. In contrast, our approach produces temporally coherent, human-like mobility patterns that consistently outperform other LLM agent baselines

\subsection{Predicting POI Popularity}

\begin{figure}[tbp]
    \centering
    \includegraphics[width=0.9\linewidth]{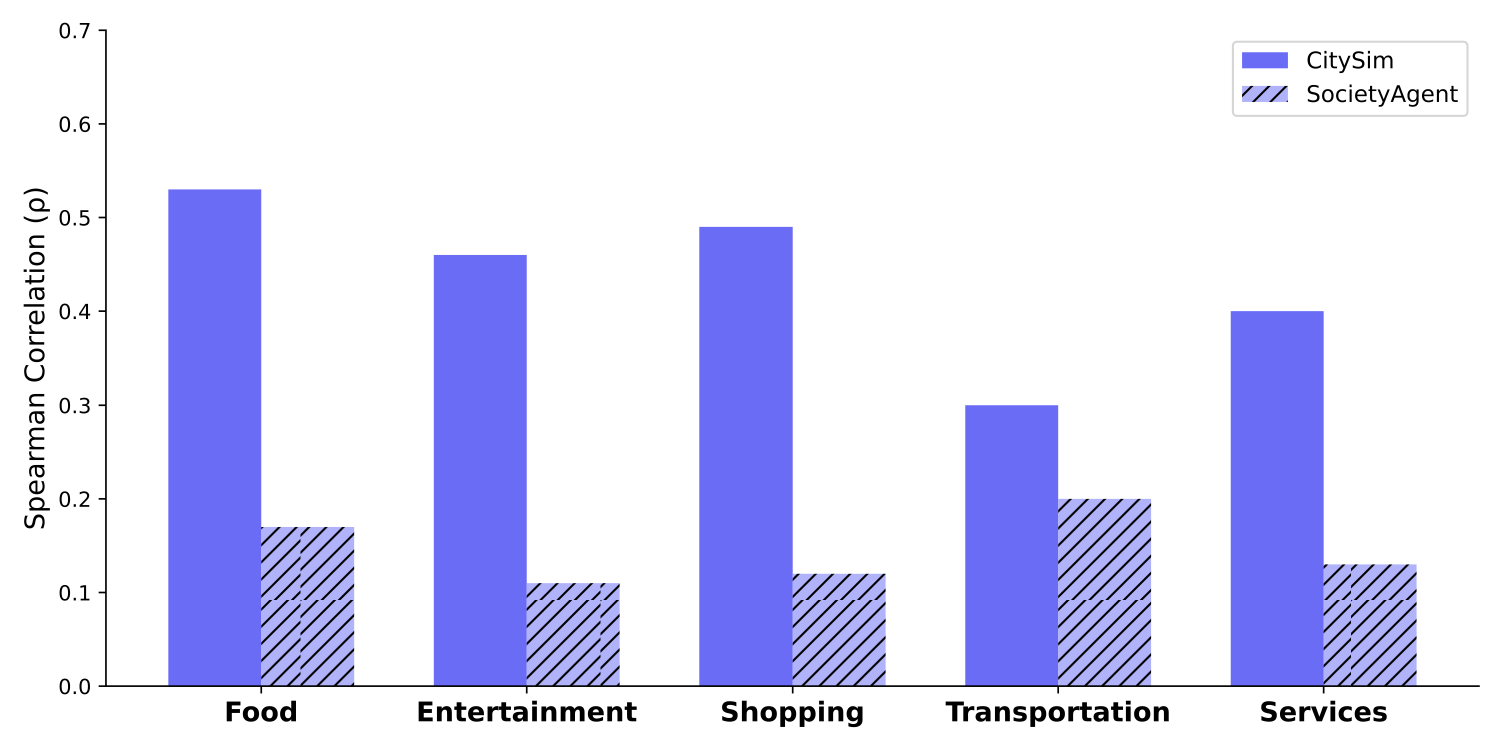}
    \caption{Comparison of real-world POI popularity and simulated-based visits in Shibuya. }
    \label{fig:poicorr}
\end{figure}
A key application is forecasting which POI will attract the most visitors, essential for urban planning, retail strategy, or event management. In light of this, we evaluate CitySim's as a predictive tool for real-world POI popularity in Shibuya (Tokyo, Japan). Ground-truth was estimated using ratings from Google Maps. Simulated popularity was measured by counting agent visits to each POI over a simulated month. We compared CitySim with SocietyAgent by calculating Spearman rank correlations between simulated and real-world popularity. As shown in Figure~\ref{fig:poicorr}, CitySim achieves positive alignment, whereas SocietyAgent yields notably weaker correlations. Notably, CitySim agents exhibit a positive bias toward well-known or branded POIs, leading to an inflated estimate of their real-world popularity. These findings show the potential of CitySim as a practical tool for predicting POI popularity for location-based business strategies.

\subsection{Social Studies using Synthetic Agents}
\begin{table}[tbp]
\centering
\small
\begin{tabular}{lcl}
\toprule
                 & F1-macro (mean $\pm$ std) &   \\
\midrule
GeAn           & 0.19 $\pm$ 0.03  & \chart{470}{32}{cyan} \\
AGA            & 0.20 $\pm$ 0.03  & \chart{485}{38}{magenta} \\
HumanoidAgent  & 0.22 $\pm$ 0.03  & \chart{550}{53}{yellow} \\
AgentSociety   & 0.28 $\pm$ 0.02  & \chart{696}{83}{magenta} \texttwemoji{3rd_place_medal} \\
MobileCity     & 0.21 $\pm$ 0.02  & \chart{522}{45}{green} \\
\rowcolor{blue!10}
CitySim   & 0.36 $\pm$ 0.02  & \chart{900}{100}{yellow} \texttwemoji{2nd_place_medal} \\
GBDT & 0.45 $\pm$ 0.04 & \chart{1120}{100}{cyan} \texttwemoji{1st_place_medal} \\
\bottomrule
\end{tabular}
\caption{
Macro F1-score for well-being class prediction (5-class) across models, evaluated on a proprietary agent survey. Medals indicate top-3 methods.
}
\label{tab:wellbeing_ablation}
\end{table}

We assess the potential of CitySim to estimate population well-being. We use a proprietary dataset comprising 1,200 well-being survey responses collected in Japan. Each record consists of a set of questions designed to estimate well-being among 5 classes. Agents were initialized with persona profiles matching the real survey respondents and engaged in three weeks of simulated city life, using their memory module to answer the same set of questions. We benchmark our method against a gradient boosting model (GBDT) trained on collected activities/locations from the dataset. As reported in Table~\ref{tab:wellbeing_ablation}, the XGBoost baseline achieves the highest macro F1-score, while CitySim closely follows and outperforms prior agent-based work. These simulation-based approaches are limited by incomplete agent background knowledge and imperfect persona initialization, which restrict their ability to fully replicate the nuances of human well-being.

\subsection{Modeling Crowd Density}

\begin{figure}[tbp]
\centering
\includegraphics[width=0.47\linewidth]{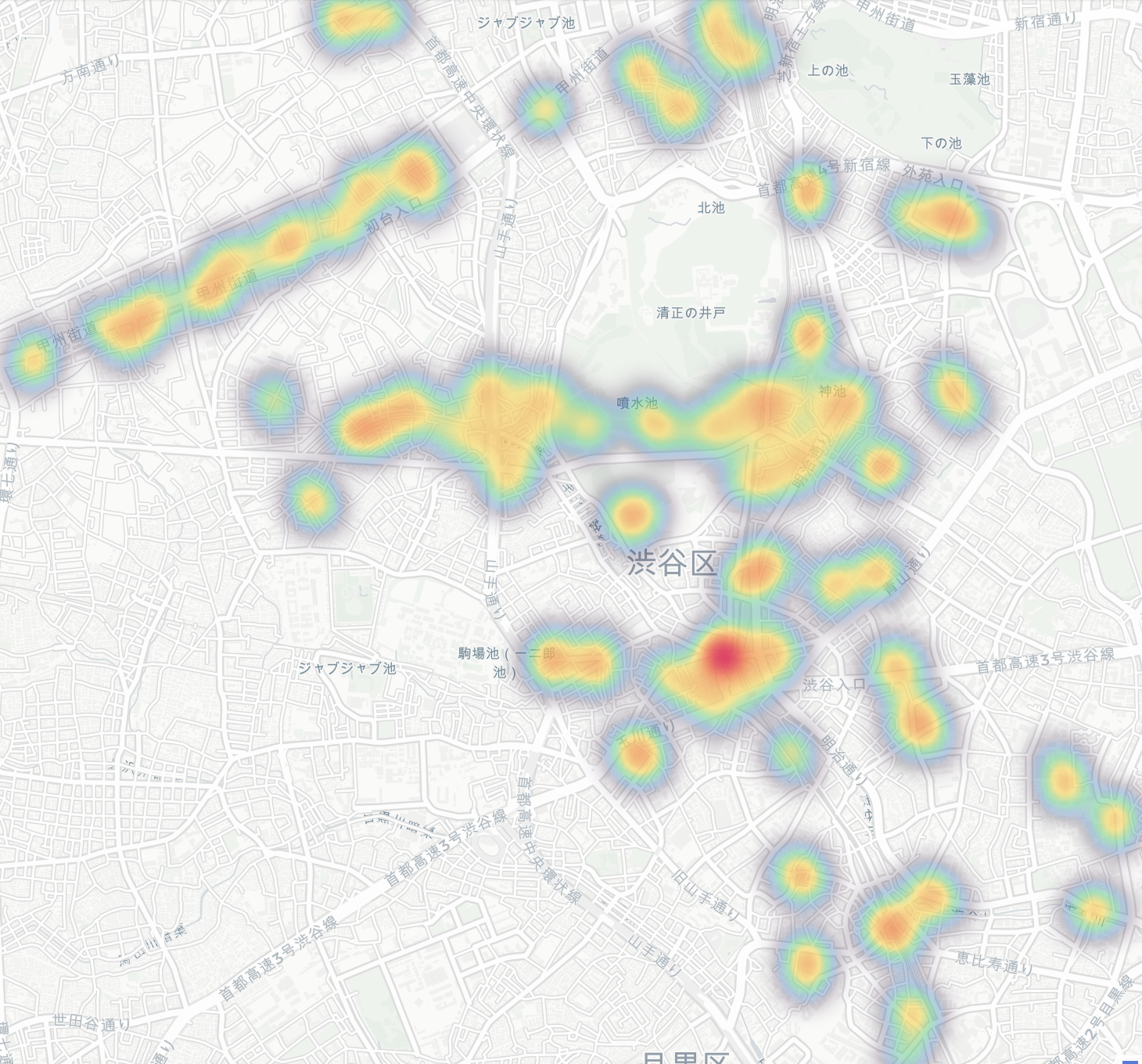}
\includegraphics[width=0.47\linewidth]{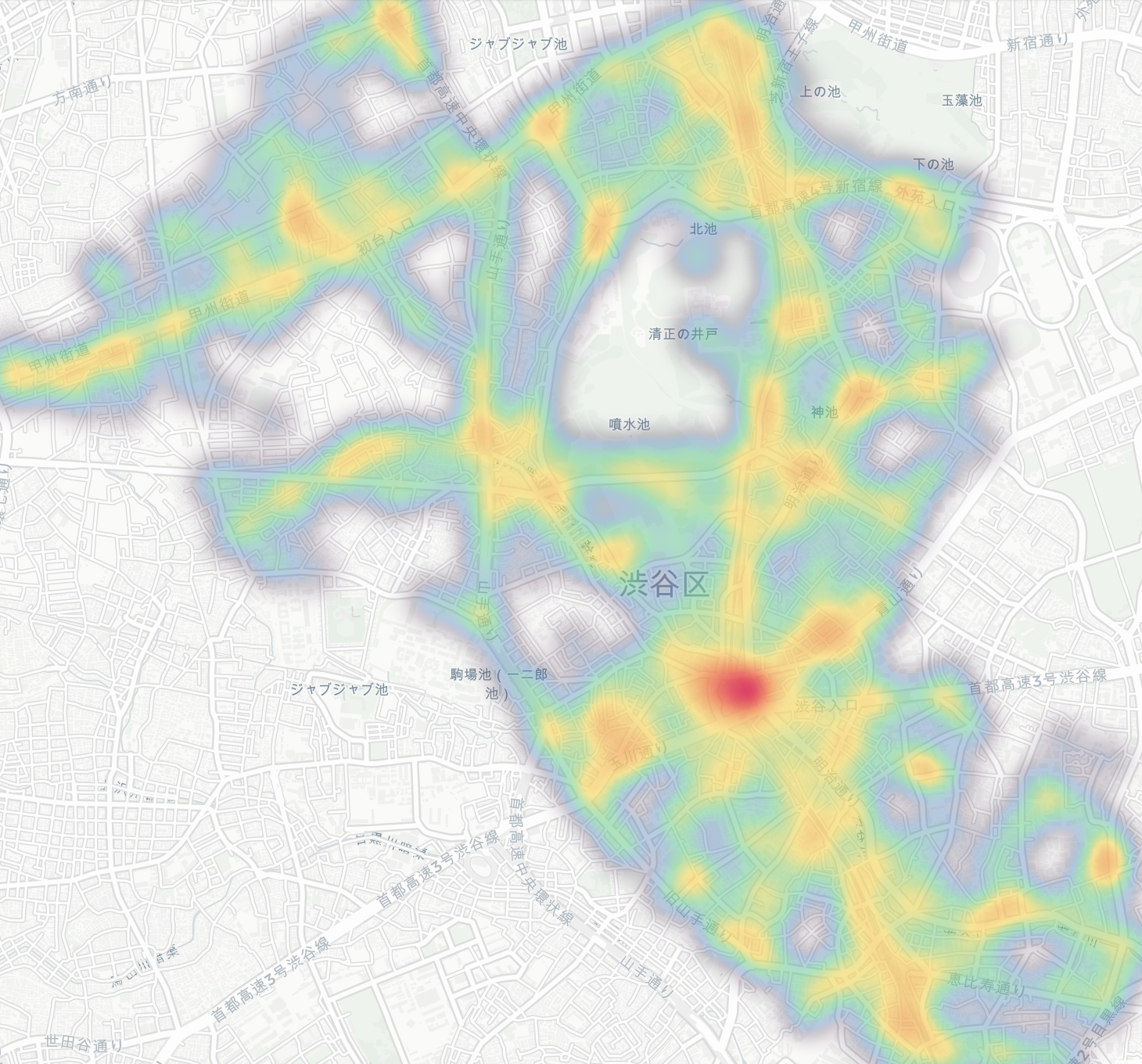}
\caption{Comparison of simulated (left) and real-world (right) crowd density heatmaps in Shibuya, Tokyo. Warmer colors indicate higher densities.}
\label{fig:density_heatmap}
\end{figure}

Predicting spatial crowd density is vital for urban management, public safety, and event planning. We now assess CitySim’s ability to reproduce real-world patterns of pedestrian concentration across Shibuya (Japan). We aggregate agent visit counts by location to generate simulated crowd density heatmaps, and compare these against ground-truth distributions estimated from smartphone location data. As shown in Figure~\ref{fig:density_heatmap}, CitySim accurately mimics mobility patterns observed in real world, with the highest densities around central transit nodes and along major commercial streets. We also notice that CitySim sometimes underestimates crowd in small streets, likely due to its belief-enhanced gravity model, which may reflect LLM popularity bias \cite{lichtenberg2024large}. This further highlights the potential of an agent-driven approach for a/b and what-if testing, including scenarios difficult to observe in real-world settings.

\section{Conclusion}
We present \textbf{CitySim}, a large-scale framework for simulating human-like urban behavior using LLM-powered agents equipped with recursive planning, real-world grounded personas, long-term goal formation, and belief-aware memory. Results demonstrate closer alignment of our agents with their human counterparts at both micro and macro levels. CitySim enables the study of complex urban phenomena and supports more realistic, adaptive agent behaviors compared to prior agent-based models. Experimental results highlight CitySim as a robust foundation for research and industry applications at the intersection of behavioral modeling and urban planning, and social studies.

\section{Limitations}
Despite achieving the best performance compared to other baselines, it is important to acknowledge several limitations of this work. A limitation of this work lies in reproducibility of this work, which is limited because the data used for some experiments is not public. Besides, our method may inherit cultural, gender, and socioeconomic biases, due to the nature of LLMs. Related to this, occasional hallucinations have been observed when generating appraisal for recent or less-known POIs, which can lead to inaccurate simulation outcomes. Moreover, the efficacy of our framework is largely reliant on the strengths and weaknesses of the underlying LLMs. The accuracy of the simulated user behavior may be impacted by LLMs' occasional inconsistent, biased, or unfounded outputs. Finally, the large number of interacting modules makes it difficult to isolate the effect of each component; we include ablation studies in the Appendix to partially address this. In the future, we will explore and improve these aspects.

\section{Ethics Statement}
This paper introduces an LLM-driven agent framework for simulating urban human behavior at scale, enabling the study of city dynamics and social behaviors in a realistic and cost-effective manner. While our approach brings clear advantages in terms of scalability and faithfulness, it also raises important ethical considerations. 

The use of synthetic agents in urban simulation may inadvertently amplify biases, such as stereotypes about age, gender, occupation, or lifestyle, if such patterns are present in the training data or agent initialization. There is also the potential risk of reinforcing or introducing new inequities in simulated urban policies, as these agents might react in ways that privilege or disadvantage certain demographic groups. Additionally, large-scale simulation of agent interactions could enable the identification and manipulation of behavioral trends, potentially informing urban interventions that steer collective behavior in subtle or non-transparent ways. This raises concerns around consent and autonomy, especially if agent outputs are used to influence real-world policies or individual choices without adequate oversight.

Finally, while synthetic agents can greatly accelerate early-stage exploration of urban scenarios, there is a risk that their deployment might marginalize the involvement of actual residents, stakeholders, and domain experts in the design and evaluation process. We recommend that synthetic humans be employed primarily to complement, not replace, human input, especially in the early phases of social studies or when involving real people poses practical or ethical challenges. By adhering to these principles, we aim to ensure that the use of synthetic urban agents is conducted in a manner that is ethical, transparent, and socially responsible.

\bibliography{custom}

\begin{thebibliography}{30}
\providecommand{\natexlab}[1]{#1}

\bibitem[{Bougie and Watanabe(2024)}]{bougie2024generative}
Nicolas Bougie and Narimasa Watanabe. 2024.
\newblock Generative adversarial reviews: When llms become the critic.
\newblock \emph{arXiv preprint arXiv:2412.10415}.

\bibitem[{Bougie and Watanabe(2025)}]{bougie2025simuser}
Nicolas Bougie and Narimasa Watanabe. 2025.
\newblock Simuser: Simulating user behavior with large language models for recommender system evaluation.
\newblock \emph{arXiv preprint arXiv:2504.12722}.

\bibitem[{Chiang and Lee(2023)}]{chiang2023can}
Cheng-Han Chiang and Hung-yi Lee. 2023.
\newblock Can large language models be an alternative to human evaluations?
\newblock \emph{arXiv preprint arXiv:2305.01937}.

\bibitem[{Epstein(1999)}]{epstein1999agent}
Joshua~M Epstein. 1999.
\newblock Agent-based computational models and generative social science.
\newblock \emph{Complexity}, 4(5):41--60.

\bibitem[{Feng et~al.(2024)Feng, Zhang, Yan, Zhang, Ouyang, Liu, Du, Guo, and Li}]{feng2024citybench}
Jie Feng, Jun Zhang, Junbo Yan, Xin Zhang, Tianjian Ouyang, Tianhui Liu, Yuwei Du, Siqi Guo, and Yong Li. 2024.
\newblock Citybench: Evaluating the capabilities of large language model as world model.
\newblock \emph{arXiv preprint arXiv:2406.13945}.

\bibitem[{Gao et~al.(2024)Gao, Lan, Li, Yuan, Ding, Zhou, Xu, and Li}]{gao2024large}
Chen Gao, Xiaochong Lan, Nian Li, Yuan Yuan, Jingtao Ding, Zhilun Zhou, Fengli Xu, and Yong Li. 2024.
\newblock Large language models empowered agent-based modeling and simulation: A survey and perspectives.
\newblock \emph{Humanities and Social Sciences Communications}, 11(1):1--24.

\bibitem[{G{\"u}rcan et~al.(2025)G{\"u}rcan, Falck, Rousseau, and Lima}]{gurcan2025towards}
{\"O}nder G{\"u}rcan, Vanja Falck, Markus~G Rousseau, and Larissa~L Lima. 2025.
\newblock Towards an llm-powered social digital twinning platform.
\newblock \emph{arXiv preprint arXiv:2505.10681}.

\bibitem[{Hofman et~al.(2021)Hofman, Watts, Athey, Garip, Griffiths, Kleinberg, Margetts, Mullainathan, Salganik, Vazire et~al.}]{hofman2021integrating}
Jake~M Hofman, Duncan~J Watts, Susan Athey, Filiz Garip, Thomas~L Griffiths, Jon Kleinberg, Helen Margetts, Sendhil Mullainathan, Matthew~J Salganik, Simine Vazire, and 1 others. 2021.
\newblock Integrating explanation and prediction in computational social science.
\newblock \emph{Nature}, 595(7866):181--188.

\bibitem[{Hong et~al.(2024)Hong, Zhuge, Chen, Zheng, Cheng, Wang, Zhang, Wang, Yau, Lin, Zhou, Ran, Xiao, Wu, and Schmidhuber}]{iclr/HongZCZCWZWYLZR24}
Sirui Hong, Mingchen Zhuge, Jonathan Chen, Xiawu Zheng, Yuheng Cheng, Jinlin Wang, Ceyao Zhang, Zili Wang, Steven Ka~Shing Yau, Zijuan Lin, Liyang Zhou, Chenyu Ran, Lingfeng Xiao, Chenglin Wu, and J{\"{u}}rgen Schmidhuber. 2024.
\newblock Metagpt: Meta programming for {A} multi-agent collaborative framework.
\newblock In \emph{International Conference on Learning Representations (ICLR)}.

\bibitem[{Huitt(2007)}]{huitt2007maslow}
William Huitt. 2007.
\newblock Maslow's hierarchy of needs.
\newblock \emph{Educational psychology interactive}, 23.

\bibitem[{LaBar and Cabeza(2006)}]{labar2006cognitive}
Kevin~S LaBar and Roberto Cabeza. 2006.
\newblock Cognitive neuroscience of emotional memory.
\newblock \emph{Nature Reviews Neuroscience}, 7(1):54--64.

\bibitem[{Lazer et~al.(2009)Lazer, Pentland, Adamic, Aral, Barab{\'a}si, Brewer, Christakis, Contractor, Fowler, Gutmann et~al.}]{lazer2009computational}
David Lazer, Alex Pentland, Lada Adamic, Sinan Aral, Albert-L{\'a}szl{\'o} Barab{\'a}si, Devon Brewer, Nicholas Christakis, Noshir Contractor, James Fowler, Myron Gutmann, and 1 others. 2009.
\newblock Computational social science.
\newblock \emph{Science}, 323(5915):721--723.

\bibitem[{Li et~al.(2023)Li, Hammoud, Itani, Khizbullin, and Ghanem}]{li2023camel}
Guohao Li, Hasan Hammoud, Hani Itani, Dmitrii Khizbullin, and Bernard Ghanem. 2023.
\newblock Camel: Communicative agents for \textquotedblleft mind\textquotedblright \ exploration of large language model society.
\newblock \emph{Advances in Neural Information Processing Systems}, 36:51991--52008.

\bibitem[{Lichtenberg et~al.(2024)Lichtenberg, Buchholz, and Schw{\"o}bel}]{lichtenberg2024large}
Jan~Malte Lichtenberg, Alexander Buchholz, and Pola Schw{\"o}bel. 2024.
\newblock Large language models as recommender systems: A study of popularity bias.
\newblock \emph{arXiv preprint arXiv:2406.01285}.

\bibitem[{López~Baeza et~al.(2021)López~Baeza, Carpio-Pinedo, Sievert, Landwehr, Preuner, Borgmann, Avakumović, Weissbach, Bruns-Berentelg, and Noennig}]{su13169268}
Jesús López~Baeza, José Carpio-Pinedo, Julia Sievert, André Landwehr, Philipp Preuner, Katharina Borgmann, Maša Avakumović, Aleksandra Weissbach, Jürgen Bruns-Berentelg, and Jörg~Rainer Noennig. 2021.
\newblock Modeling pedestrian flows: Agent-based simulations of pedestrian activity for land use distributions in urban developments.
\newblock \emph{Sustainability}, 13(16).

\bibitem[{Macal and North(2005)}]{macal2005tutorial}
Charles~M Macal and Michael~J North. 2005.
\newblock Tutorial on agent-based modeling and simulation.
\newblock In \emph{Proceedings of the Winter Simulation Conference, 2005.}, pages 14--pp. IEEE.

\bibitem[{{OpenStreetMap contributors}(2025)}]{openstreetmap}
{OpenStreetMap contributors}. 2025.
\newblock \href {https://www.openstreetmap.org} {Openstreetmap}.

\bibitem[{Park et~al.(2023{\natexlab{a}})Park, O'Brien, Cai, Morris, Liang, and Bernstein}]{park2023generative}
Joon~Sung Park, Joseph O'Brien, Carrie~Jun Cai, Meredith~Ringel Morris, Percy Liang, and Michael~S Bernstein. 2023{\natexlab{a}}.
\newblock Generative agents: Interactive simulacra of human behavior.
\newblock In \emph{Proceedings of the 36th Annual ACM Symposium on User Interface Software and Technology}, pages 1--22.

\bibitem[{Park et~al.(2023{\natexlab{b}})Park, O'Brien, Cai, Morris, Liang, and Bernstein}]{uist/ParkOCMLB23}
Joon~Sung Park, Joseph~C. O'Brien, Carrie~Jun Cai, Meredith~Ringel Morris, Percy Liang, and Michael~S. Bernstein. 2023{\natexlab{b}}.
\newblock Generative agents: Interactive simulacra of human behavior.
\newblock In \emph{The 36th Annual Symposium on User Interface Software and Technology (UIST)}, pages 2:1--2:22.

\bibitem[{Park et~al.(2024)Park, Zou, Shaw, Hill, Cai, Morris, Willer, Liang, and Bernstein}]{corr/abs-2411-10109}
Joon~Sung Park, Carolyn~Q. Zou, Aaron Shaw, Benjamin~Mako Hill, Carrie~J. Cai, Meredith~Ringel Morris, Robb Willer, Percy Liang, and Michael~S. Bernstein. 2024.
\newblock \href {https://doi.org/10.48550/ARXIV.2411.10109} {Generative agent simulations of 1,000 people}.
\newblock \emph{CoRR}, abs/2411.10109.

\bibitem[{Piao et~al.(2025)Piao, Yan, Zhang, Li, Yan, Lan, Lu, Zheng, Wang, Zhou et~al.}]{piao2025agentsociety}
Jinghua Piao, Yuwei Yan, Jun Zhang, Nian Li, Junbo Yan, Xiaochong Lan, Zhihong Lu, Zhiheng Zheng, Jing~Yi Wang, Di~Zhou, and 1 others. 2025.
\newblock Agentsociety: Large-scale simulation of llm-driven generative agents advances understanding of human behaviors and society.
\newblock \emph{arXiv preprint arXiv:2502.08691}.

\bibitem[{{Statistics Bureau of Japan}(2021)}]{e-stat2021timeuse}
{Statistics Bureau of Japan}. 2021.
\newblock Survey on time use and leisure activities, 2021.
\newblock \url{https://www.e-stat.go.jp/en/stat-search/files?page=1&toukei=00200533&tstat=000001158160}.
\newblock Accessed July 2024.

\bibitem[{Wang et~al.(2023)Wang, Chiu, and Chiu}]{emnlp/WangCC23}
Zhilin Wang, Yu{-}Ying Chiu, and Yu~Cheung Chiu. 2023.
\newblock Humanoid agents: Platform for simulating human-like generative agents.
\newblock In \emph{Conference on Empirical Methods in Natural Language Processing (EMNLP)}.

\bibitem[{Wei et~al.(2022)Wei, Wang, Schuurmans, Bosma, Xia, Chi, Le, Zhou et~al.}]{wei2022chain}
Jason Wei, Xuezhi Wang, Dale Schuurmans, Maarten Bosma, Fei Xia, Ed~Chi, Quoc~V Le, Denny Zhou, and 1 others. 2022.
\newblock Chain-of-thought prompting elicits reasoning in large language models.
\newblock \emph{Advances in neural information processing systems}, 35:24824--24837.

\bibitem[{Wilensky(2015)}]{wilensky2015introduction}
U~Wilensky. 2015.
\newblock \emph{An Introduction to Agent-Based Modeling: Modeling Natural, Social, and Engineered Complex Systems with Netlogo}.
\newblock The MIT Press.

\bibitem[{Xia et~al.(2018)Xia, Wang, Kong, Wang, Li, and Liu}]{xia2018exploring}
Feng Xia, Jinzhong Wang, Xiangjie Kong, Zhibo Wang, Jianxin Li, and Chengfei Liu. 2018.
\newblock Exploring human mobility patterns in urban scenarios: A trajectory data perspective.
\newblock \emph{IEEE Communications Magazine}, 56(3):142--149.

\bibitem[{Ye et~al.(2025)Ye, Bougie, Yamasaki, and Watanabe}]{ye2025mobilecity}
Xiaotong Ye, Nicolas Bougie, Toshihiko Yamasaki, and Narimasa Watanabe. 2025.
\newblock Mobilecity: An efficient framework for large-scale urban behavior simulation.
\newblock \emph{arXiv preprint arXiv:2504.16946}.

\bibitem[{Yu et~al.(2024)Yu, Zhang, Li, Fu, and Ye}]{corr/abs-2402-02053}
Yangbin Yu, Qin Zhang, Junyou Li, Qiang Fu, and Deheng Ye. 2024.
\newblock \href {https://doi.org/10.48550/ARXIV.2402.02053} {Affordable generative agents}.
\newblock \emph{CoRR}, abs/2402.02053.

\bibitem[{Zheng et~al.(2022)Zheng, Trott, Srinivasa, Parkes, and Socher}]{zheng2022ai}
Stephan Zheng, Alexander Trott, Sunil Srinivasa, David~C Parkes, and Richard Socher. 2022.
\newblock The ai economist: Taxation policy design via two-level deep multiagent reinforcement learning.
\newblock \emph{Science advances}, 8(18):eabk2607.

\bibitem[{Zhou et~al.(2023)Zhou, Jiang, Li, Wu, Wang, Qiu, Zhang, Chen, Wu, Wang, Zhu, Chen, Zhang, Zhang, Chen, Cui, and Sachan}]{corr/abs-2309-07870}
Wangchunshu Zhou, Yuchen~Eleanor Jiang, Long Li, Jialong Wu, Tiannan Wang, Shi Qiu, Jintian Zhang, Jing Chen, Ruipu Wu, Shuai Wang, Shiding Zhu, Jiyu Chen, Wentao Zhang, Ningyu Zhang, Huajun Chen, Peng Cui, and Mrinmaya Sachan. 2023.
\newblock \href {https://doi.org/10.48550/ARXIV.2309.07870} {Agents: An open-source framework for autonomous language agents}.
\newblock \emph{CoRR}, abs/2309.07870.

\end{thebibliography}
\clearpage
\appendix

\begin{figure}[tbp]
    \centering
    \includegraphics[width=1.0\linewidth]{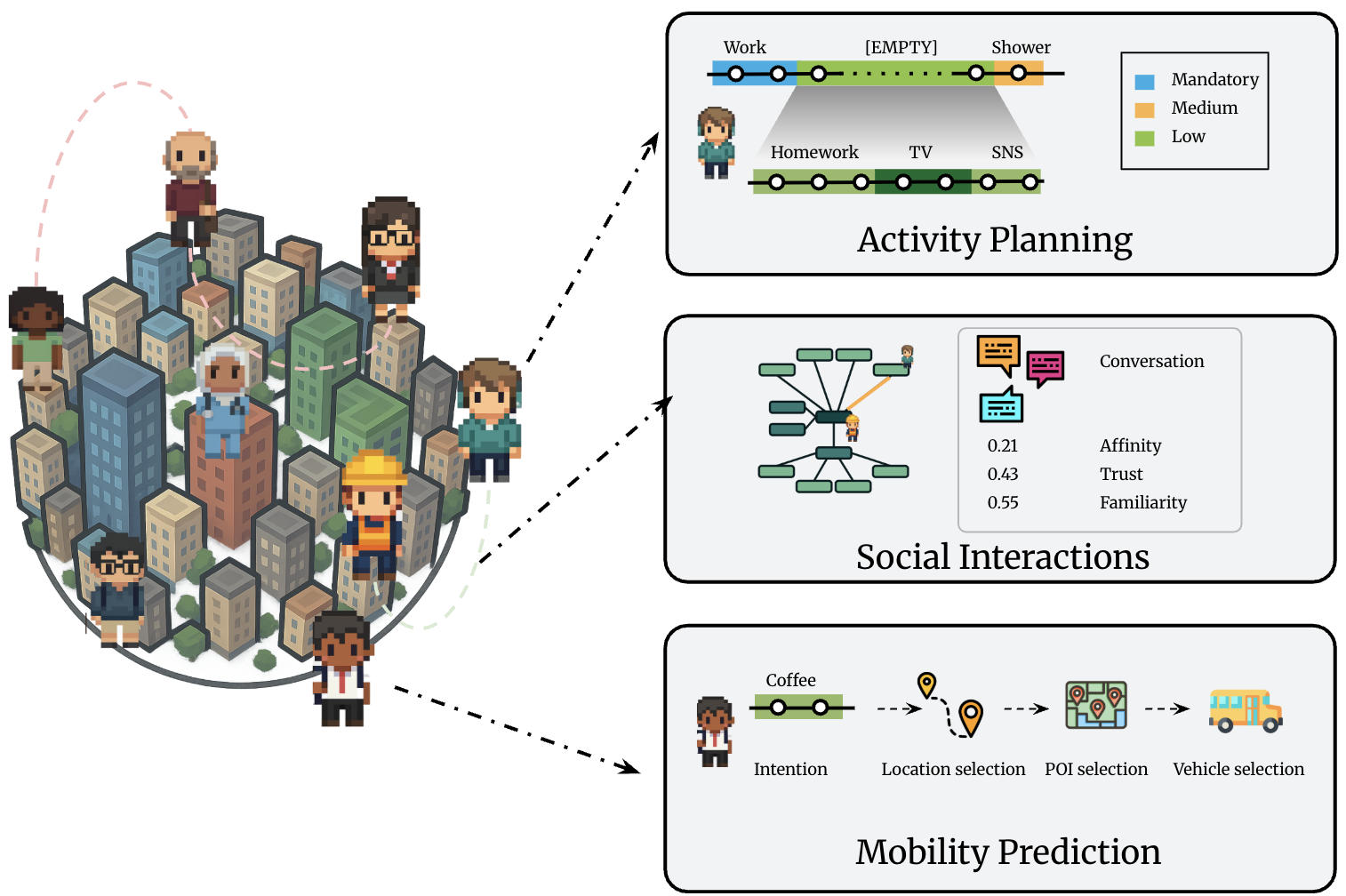}
    \caption{Overview of CitySim: LLM-based agents with diverse personas plan daily activities, interact socially, and navigate a virtual city environment.
    }
    \label{fig:overall_method}
\end{figure}

\sethlcolor{customblue}

\section{Experimental Setup}
We provide an illustration of CitySim in Figure \ref{fig:overall_method}. All agent attributes in the persona module are initialized from a proprietary survey-based dataset, conducted in Japan. The attribute distributions closely match those observed in recent Japanese census statistics and lifestyle surveys \cite{e-stat2021timeuse}. Big Five personality traits are discretized on a 3-point scale (1=low, 2=medium, 3=high). Each agent’s home and workplace (or school) locations are assigned according to Japanese population density, based on OpenStreetMap data \cite{openstreetmap}, ensuring realistic urban spatial distribution and feasible commutes. 

The temporal memory retrieves the top $k_1=5$ entries from the past $\Delta t=24$ hours (cosine embedding similarity); each memory node stores \{time, location, observation, key\}. Reflective memory is synthesized at the end of each day, linking to temporal events. Spatial memory beliefs $b_i \in \mathbb{R}^4$ (\{price, atmosphere, satisfaction, convenience\}) are initialized as $0.5$ (neutral) for unvisited POIs, updated at each visit via a Kalman filter, acknowledging this slight abuse of notation for convenience:
\begin{equation}
\resizebox{\linewidth}{!}{$
\begin{aligned}
K^{(t)} &= \frac{\sigma_{b}^{(t-1)}}{\sigma_{b}^{(t-1)} + \sigma_{o}} \quad
b_{i}^{(t)} = K^{(t)} o_{i}^{(t)} + (1 - K^{(t)}) b_{i}^{(t-1)} \\
\sigma_{b}^{(t)} &= (1 - K^{(t)}) \sigma_{b}^{(t-1)}
\end{aligned}
$}
\end{equation} 
where $\sigma_{b}^{(t)}$ denotes uncertainties in beliefs, and with $\sigma_b^{(0)}=0.25$ and $\sigma_o=0.2$, and subject to daily decay with $\lambda=0.03$. Beliefs for unvisited POIs are imputed from the $k=10$ most similar visited locations based on embedding distances. 

Need thresholds for action interruption are $T_{\text{hunger}}=0.3$, $T_{\text{energy}}=0.3$, $T_{\text{safety}}=0.2$, $T_{\text{social}}=0.2$, and priority order is \textit{hunger} $>$ \textit{safety} $>$ \textit{energy} $>$ \textit{social}. Long-term goals are revised monthly or after major life events, using a need fulfillment index (representing the proportion of the day when all core needs exceed their thresholds), financial stress (income $<$ 0.9$\times$expenses), and social isolation (fewer than 3 unique contacts in 7 days) as triggers. Simulation operates with a 5 minute timestep, and all random seeds are fixed for reproducibility.

Daily schedules are constructed from time blocks with a minimum granularity of 5 minutes, matching the resolution of human routine reporting in time-use surveys. In value-driven planning, $N=3$ candidate activities are generated for each leisure block, balancing agent diversity and computational efficiency. During place selection, the agent considers the top 10 nearby areas and up to 200 candidate POIs by relevance and proximity for each activity. The gravity model uses $\gamma=2.0$ as distance decay, with $\varepsilon=10^{-3}$ for stability. Belief score considered in the gravity model is the the average of the agent’s beliefs about this place across four dimensions, as recorded in its spatial memory. For transport, available vehicles include walk, bicycle, car, bus, and train.  Face-to-face interactions are limited to one partner per 30-minute tick to avoid excessive social behaviors.

At the end of each day, the agent synthesizes higher-level insights regarding its habits, preferences, and beliefs through a structured reflection process. Specifically, the most recent entries from the agent’s temporal memory are compiled and provided to the language model. The initial prompt instructs the model to identify salient, high-level questions that can be answered using only the provided memory records. The questions in our approach are explicitly formulated to uncover recurring patterns of behavior (habits), preference, and evolving beliefs. For each generated question, relevant memory entries are retrieved using semantic similarity and temporal proximity. The language model is then prompted to extract up to five distinct insights, each supported by explicit references to the underlying memory entries (e.g., \texttt{The agent prefers evening study sessions (evidence: 8, 13, 22, 45)}). Both direct observations and prior reflective statements are eligible to serve as evidence, enabling the recursive construction of abstracted self-knowledge. All resulting insights are stored in the reflective memory.

For pairwise human preference, we defined the following criterion: (i) \textbf{Naturalness}—the extent to which actions align with the agent’s profile, habits, and context; (ii) \textbf{Coherence}—the logical progression and goal-directedness of activities; and (iii) \textbf{Plausibility}—the overall believability of the sequence given realistic urban behavior.

\subsection{Module Details}

We now provide a comprehensive explanation of some modules, detailing the implementation and technical details.

\subsubsection{Planning Module}
Daily planning follows a two-step recursive approach:
\begin{enumerate}
    \item \textbf{Mandatory Block Assignment:} Starting from an empty day, the planner assigns fixed, non-negotiable activities (e.g., sleep, work, medical appointments) using agent persona, occupation, and needs. If a selected activity does not fill the entire interval, the block may be subdivided according to the activity duration.
    \item \textbf{Medium-Priority Recursive Filling:}
    After planning mandatory activities, remaining \texttt{[EMPTY]} blocks are recursively processed for medium-priority tasks (e.g., meals, hygiene, essential errands). 
\end{enumerate}

\noindent
One may notice that some \texttt{[blocks]} remain empty. Following Maslow’s Hierarchy of Needs \cite{huitt2007maslow}, those blocks are typically reserved for leisure activities or to satisfy long-term goals/needs (e.g., hobbies, socializing, exploration). Therefore, they are filled at execution based on the agent's current state, location, dominant needs, future schedule, using a \textbf{value-driven planning}. For each empty interval, we argue that the paradigm of presenting multiple candidate activities and evaluating them enables the agent to select the best action. Thus, we prompt the agent to generate $N$ candidates --- with maximum duration being the block, that may improve enjoyment, satisfaction, or fulfill a need or goal. Next, it evaluates each activity by imagining the resulting desire states that the agent would experience after taking action. Based on these evaluations, the agent selects the activity at that is expected to best fulfill the agent’s intrinsic desires. This scheme is executed through a single structured LLM call with multiple internal reasoning steps.

\subsubsection{Social Module}
Beyond maintaining evolving belief vectors over social ties, our social module incorporates structured reasoning to support context-sensitive communication. When initiating a conversation, agents rely on LLM-generated judgments that consider not only relationship strength but also the agent's intention, emotional state, and ongoing thought processes. Message generation is similarly guided by a prompt framework that reflects personality traits, past interactions, and constrained discussion topics derived from the agent’s persona. 

Following each interaction, beliefs are updated by evaluating the sentiment and outcome of the exchange: positive, neutral, or negative signals are extracted from the conversation and used to incrementally adjust the affinity, trust, and familiarity scores between agents. In addition, social interactions are not statically scheduled. Instead, unmet social needs dynamically trigger acquaintance search and interaction planning. For instance, when social satisfaction falls below a threshold, agents proactively evaluate whom to contact and whether the mode of interaction should be face-to-face or online. This decision process is handled by a single LLM call, producing both the modality and the target agent in structured form. These features enable the emergence of diverse, adaptive social patterns across agents over time.

\section{Discussion}
We acknowledge that our method exhibits certain limitations. The collective behaviors generated by CitySim agents are well-aligned with established theories in urban studies and commonly observed patterns in city life. Micro-level phenomena, such as individual activity selection, place visits, and route planning, emerge from the endogenous decision-making of our agents. However, the underlying reasons for why agents exhibit specific motivational and planning patterns remain partially unexplained due to the inherent black-box nature of large language models. One possible explanation is that LLMs encode knowledge and behaviors present in their diverse training corpora, which includes textual descriptions of urban routines, spatial preferences, and daily life across global contexts.

A further limitation stems from the dependency on sufficient behavioral and interactional data to construct detailed and faithful agent personas. In some cases, real-world data may be limited, particularly for cold-start populations or marginalized user groups with fewer observed interactions. This constraint reduces the effectiveness of modules such as persona, belief, or long-term goal formation, which rely on rich historical context. To mitigate this, we initialize agent personas using a diverse set of demographic features (age, occupation, life stage) and personality traits sampled from empirical distributions, but this remains an imperfect proxy of geniune humans.

As with any LLM-based simulation, there is a risk that model-driven agents inherit biases present in large-scale data, potentially leading to the underrepresentation or oversimplification of certain urban groups or behaviors. This can pose challenges when applying simulation results to real-world urban policy or planning, as the needs of underrepresented groups might be overlooked. To address this, our experiments ensure a broad spectrum of simulated personas --- encompassing occupations, age groups, and personality profiles, and we quantify discrepancies between generated and real-world behavior distributions. As future work, we aim to analyze the representation of minority and vulnerable groups in our synthetic society, and to extend CitySim to additional domains (e.g., health, mobility, food environments).

Some experiments in our paper rely on LLM-as-judge evaluations using GPT-4o, while GPT-4o-mini powers the agents themselves. Although this circular evaluation approach may introduce significant bias, as LLMs tend to favor content generated in their own style, it remains a common practice due to the scalability and consistency offered by automated evaluations. Nevertheless, we acknowledge the limitations inherent in this methodology, including the potential for inflated performance metrics and diminished generalizability to real human judgment. To mitigate these concerns, future work should incorporate more diverse evaluation strategies, including human assessments and cross-model validation.

While CitySim accurately reproduces major crowd patterns in central and highly accessible areas, it tends to underrepresent pedestrian density in smaller streets due to the belief-enhanced gravity model’s emphasis on prominent POIs. To mitigate this limitation, future work could integrate additional urban context features—such as land use data, pedestrian infrastructure, or historical mobility traces—into the location selection process, enabling agents to account for micro-scale attractors and local accessibility \cite{xia2018exploring,su13169268}. Moreover, incorporating adaptive behavioral modules that encourage agents to occasionally explore less popular areas, either through learned novelty-seeking or routine variation, may help better capture the diversity of real-world movement \cite{gurcan2025towards}.

Finally, while CitySim faithfully models daily routines, belief formation, and adaptive planning, it abstracts away some contextual factors that influence real-world appraisal, including weather conditions, crowding, transportation delays, or accessibility constraints. Moreover, some internal needs like self-esteem and self-actualization are not yet fully represented, as they are subjective and often depend on individual values, goals, and life circumstances. This introduces a potential gap between the richness of real urban interactions and our agent-based simulation. Capturing these factors requires more nuanced modeling beyond observable behaviors, which presents ongoing challenges for simulation environments like CitySim.

\subsection{Pseudo-Code}
We provide the pseudo-code of our method.
\begin{tcolorbox}[
    colback=citysimlight,
    colframe=citysimaccent,
    boxrule=0.9pt,
    sharp corners=south,
    title=\textbf{Algorithm 1: Daily Simulation Loop},
    fonttitle=\bfseries,
    left=2mm, right=2mm, top=1mm, bottom=1mm
]
\small
\textbf{For each day:}
\vspace{0.2em}

\hspace*{1.2em}\textbf{For each agent:}
\vspace{0.1em}

\hspace*{2.2em}\texttt{plan\_day()}
\vspace{0.1em}

\hspace*{2.2em}\textbf{For each time step:}
\vspace{0.1em}

\hspace*{3.2em}\texttt{perceive()}
\vspace{0.1em}

\hspace*{3.2em}action $\leftarrow$ \texttt{decide\_action()}
\vspace{0.1em}

\hspace*{3.2em}\textbf{if} action.requires\_move:
\vspace{0.1em}

\hspace*{4.2em}poi $\leftarrow$ \texttt{select\_POI()}
\vspace{0.1em}

\hspace*{4.2em}vehicle $\leftarrow$ \texttt{select\_vehicle(poi)}
\vspace{0.1em}

\hspace*{4.2em}\texttt{move(poi, vehicle)}
\vspace{0.1em}

\hspace*{3.2em}\textbf{else:}
\vspace{0.1em}

\hspace*{4.2em}\texttt{execute(action)}
\vspace{0.1em}

\hspace*{3.2em}\texttt{reflect()} \hspace{0.5em} \textit{//beliefs, goals, needs, habits, ...}
\end{tcolorbox}

\section{Additional Experiments}

\subsection{Performance Evaluation}
\begin{table}[tbp]
    \centering
    \resizebox{0.98\linewidth}{!}{
    \begin{tabular}{lcc}
    \toprule
    \# Agents &
    \textbf{CitySim} (mean ± SD) [s] &
    \textbf{AgentSociety} (mean ± SD) [s] \\
    \midrule
    $10^3$   & $9.0 \times 10^{-3} \pm 3.2 \times 10^{-5}$   & $8.6 \times 10^{-3} \pm 3.0 \times 10^{-5}$ \\
    $10^4$   & $9.7 \times 10^{-3} \pm 2.1 \times 10^{-5}$   & $9.1 \times 10^{-3} \pm 1.5 \times 10^{-5}$ \\
    $10^5$   & $2.1 \times 10^{-2} \pm 5.0 \times 10^{-4}$   & $1.8 \times 10^{-2} \pm 5.7 \times 10^{-4}$ \\
    $10^6$   & $0.183 \pm 5.6 \times 10^{-4}$                & $0.168 \pm 5.3 \times 10^{-4}$ \\
    \bottomrule
    \end{tabular}}
    \caption{Mean time per simulation step (seconds) and standard deviation as a function of agent population. }
    \label{tab:perf_comparison}
\end{table}
To assess the scalability and efficiency of our city simulation framework, we conduct a performance analysis. We simulate agent populations of $10^3$, $10^4$, $10^5$, and $10^6$ individuals, distributing their departure times according to typical weekday peaks. As done in AgentSociety \cite{piao2025agentsociety}, for each agent, we alternate between setting and fetching queries (at a 1:999 ratio) to mimic realistic simulation workloads. The simulation runs for 24 virtual hours, with the main metric being the mean simulation step time per agent. Each setting is repeated five times to obtain average and standard deviation values. The simulation speed results are presented in Table~\ref{tab:perf_comparison}. We observe that, even as both the agent population and query frequency grow by several orders of magnitude, the average time per simulation step increases modestly. This demonstrates that our framework supports large-scale simulations with negligible loss of efficiency, and is well-suited for modeling complex urban scenarios with massive agent populations.

\subsection{Human Likeliness}
\begin{table}[btp]
\centering
\resizebox{1.0\columnwidth}{!}{
\begin{tabular}{lcccc}
\toprule
\textbf{Method} & \textbf{Activity} & \textbf{Dialogue} & \textbf{Mobility} & \textbf{Event Reaction} \\
\midrule
GeAn & 3.11 $\pm$ 0.18 & 3.96 $\pm$ 0.04 & 3.08 $\pm$ 0.17 & 3.03 $\pm$ 0.21 \\
AGA & 3.22 $\pm$ 0.28 & 4.00 $\pm$ 0.03 & 3.16 $\pm$ 0.24 & 3.15 $\pm$ 0.19 \\
HumanoidAgent & 3.30 $\pm$ 0.31 & 3.99 $\pm$ 0.05 & 3.29 $\pm$ 0.22 & 3.21 $\pm$ 0.17 \\
AgentSociety & 4.02 $\pm$ 0.22 & 4.08 $\pm$ 0.06 & 3.82 $\pm$ 0.25 & 3.75 $\pm$ 0.21 \\
MobileCity & 4.09 $\pm$ 0.27 & 4.04 $\pm$ 0.06 & 3.96 $\pm$ 0.18 & 3.89 $\pm$ 0.17 \\
\rowcolor{blue!10}
CitySim & \textbf{4.37} $\pm$ \textbf{0.18} & \textbf{4.23} $\pm$ \textbf{0.04} & \textbf{4.14} $\pm$ \textbf{0.15} & \textbf{4.09} $\pm$ \textbf{0.16} \\
\bottomrule
\end{tabular}}
\caption{Human-likeness score evaluated by GPT-4o across city simulation domains. Higher values indicate greater similarity to real human responses.}
\label{tab:llm_humanlike_city}
\end{table}

As LLM Evaluators \cite{chiang2023can} have demonstrated performance on par with human annotators, we leverage GPT-4o to judge whether agent behaviors in our simulation appear human or LLM generated. For each baseline, we collect 20,000 outputs across four domains: daily activities, dialogue, mobility choices, and event reactions. GPT-4o assesses each sample using a 5-point Likert scale, with higher scores indicating stronger resemblance to human-like responses. Results in Table~\ref{tab:llm_humanlike_city} show that our model significantly outperforms all baselines across the evaluated domains. Notably, the integration of value-driven planning and belief-aware mobility contributes to substantial improvements in both the mobility and event reaction scores. Furthermore, our agent's explicit modeling of needs, feelings, and long-term goals leads to more consistent and believable routines. In contrast, baseline agents are more likely to produce repetitive actions or unrealistic schedules --- unusually long breakfasts or work starting atypically late, contributing to suspicions of AI involvement.

\subsection{Needs Evolution}
\begin{figure}[tbp]
    \centering
    \includegraphics[width=1.0\linewidth]{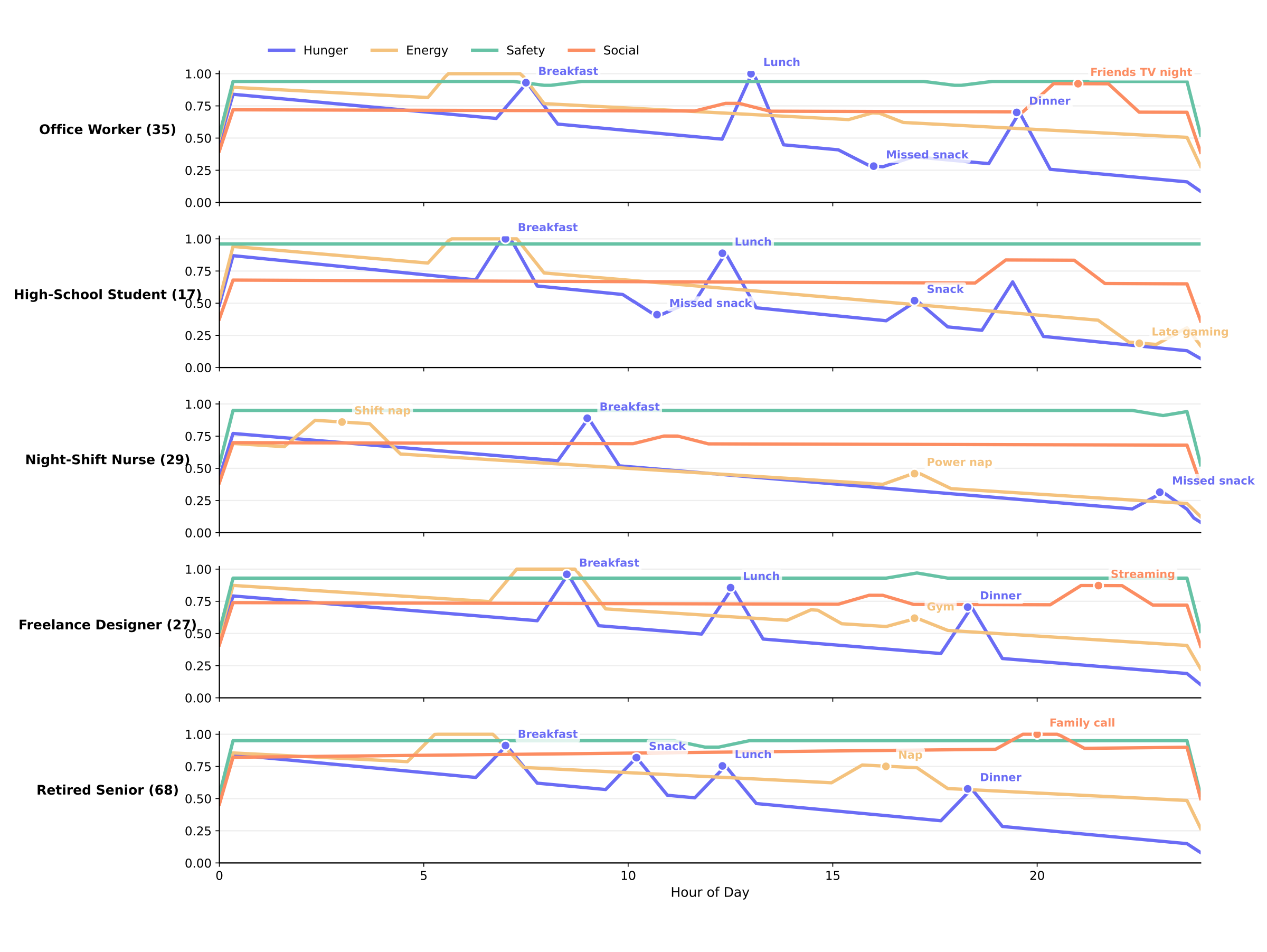}
    \caption{Evolution of basic needs (hunger, energy, safety, social satisfaction) for five agents across one day.}
    \label{fig:needs}
\end{figure}
To assess the realism and adaptability of our agent-based planning framework, we simulate the evolution of basic needs across a typical day for five agent profiles: a mid-career office worker, a high-school student, a night-shift nurse, a freelance designer, and a retired senior. Figure~\ref{fig:needs} presents the resulting need trajectories, which display distinct, context-dependent patterns consistent with real-world routines. The office worker and student exhibit dips in hunger and energy preceding following periods of sustained activity, with occasional snacks to restore hunger during the day. The night-shift nurse’s trajectories capture irregular sleep and meal patterns inherent to shift work, while the freelance designer demonstrates variable self-care and flexible scheduling. The retired senior shows more regular mealtimes, napping, and consistently high social satisfaction due to living with their family. 

\begin{table*}[tbp]
\centering
\small
\setlength{\tabcolsep}{4.5pt}
\begin{tabular}{lclccclccclccclcc}
\toprule
                 & \multicolumn{4}{c}{Activity} & \multicolumn{4}{c}{Dialogue} & \multicolumn{4}{c}{Mobility} & \multicolumn{4}{c}{Event Reaction} \\
\cmidrule(lr){2-5} \cmidrule(lr){6-9} \cmidrule(lr){10-13} \cmidrule(lr){14-17}
                 & Value &  &  &  & Value &  &  &  & Value &  &  &  & Value &  &  &  \\
\midrule
\rowcolor{blue!10}
CitySim (full)      & 4.37 $\pm$ 0.18 & \chart{1190}{100}{cyan} & \texttwemoji{1st_place_medal} & 
                & 4.23 $\pm$ 0.04 & \chart{1174}{96}{magenta} & \texttwemoji{1st_place_medal} & 
                & 4.14 $\pm$ 0.15 & \chart{1162}{93}{yellow} & \texttwemoji{1st_place_medal} & 
                & 4.09 $\pm$ 0.16 & \chart{1144}{85}{green} & \texttwemoji{1st_place_medal} \\
w/o Belief      & 3.85 $\pm$ 0.22 & \chart{1069}{66}{cyan} & \texttwemoji{2nd_place_medal} & 
                & 3.92 $\pm$ 0.08 & \chart{1095}{74}{magenta} & \texttwemoji{2nd_place_medal} & 
                & 3.75 $\pm$ 0.18 & \chart{1031}{56}{yellow} & \texttwemoji{2nd_place_medal} & 
                & 3.60 $\pm$ 0.21 & \chart{975}{41}{green} & \texttwemoji{2nd_place_medal} \\
w/o Rec. Plan   & 3.72 $\pm$ 0.17 & \chart{1026}{43}{cyan} &  &  
                & 3.85 $\pm$ 0.06 & \chart{1073}{62}{magenta} &  &  
                & 3.80 $\pm$ 0.17 & \chart{1046}{62}{yellow} &  &  
                & 3.65 $\pm$ 0.16 & \chart{988}{44}{green} &  &  \\
w/o LT Goal     & 3.80 $\pm$ 0.19 & \chart{1047}{53}{cyan} &  &  
                & 3.95 $\pm$ 0.07 & \chart{1104}{79}{magenta} & \texttwemoji{3rd_place_medal}  &  
                & 3.88 $\pm$ 0.14 & \chart{1070}{75}{yellow} & \texttwemoji{3rd_place_medal}  &  
                & 3.70 $\pm$ 0.18 & \chart{1001}{47}{green} & & \\
w/o Needs       & 3.55 $\pm$ 0.21 & \chart{988}{44}{cyan} &  &  
                & 3.82 $\pm$ 0.09 & \chart{1064}{58}{magenta} &  &  
                & 3.73 $\pm$ 0.16 & \chart{1025}{51}{yellow} &  &  
                & 3.50 $\pm$ 0.20 & \chart{950}{29}{green} &  & \\
w/o Persona     & 3.60 $\pm$ 0.20 & \chart{1001}{49}{cyan} &  &  
                & 3.60 $\pm$ 0.10 & \chart{1001}{49}{magenta} &  &  
                & 3.72 $\pm$ 0.17 & \chart{1021}{50}{yellow} &  &  
                & 3.58 $\pm$ 0.17 & \chart{970}{38}{green} &  & \\
\bottomrule
\end{tabular}
\caption{
Ablation study for CitySim, evaluated by GPT-4o. Metrics reflect human-likeness (Likert, 1-5 scale, mean $\pm$ std) for activity, dialogue, mobility, and event reaction. Medals denote top-3 performance in each domain.
}
\label{tab:ablation_citysim}
\end{table*}

\subsection{Belief Estimation}

\begin{figure*}[tbp]
    \centering
    \includegraphics[width=1.0\linewidth]{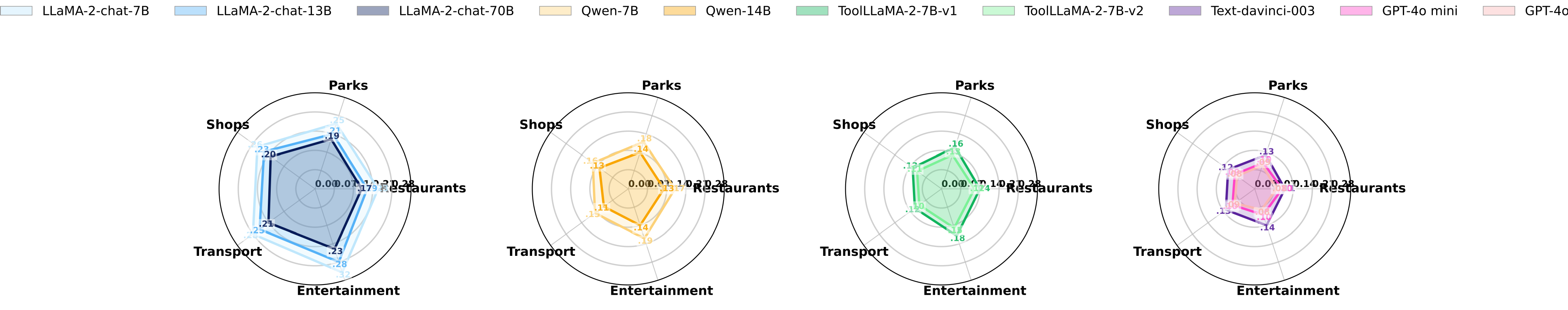}
    \caption{Category-wise mean absolute error (MAE) of belief estimation for unvisited POIs, evaluated across five semantic categories (\emph{Restaurants}, \emph{Parks}, \emph{Shops}, \emph{Transport}, \emph{Entertainment}) and ten LLM-based agent models. Lower values indicate higher accuracy. }
    \label{fig:belief}
\end{figure*}
Beliefs are central in shaping human behaviors. To evaluate the accuracy and consistency of our belief estimation, we conduct a study in which each agent is first initialized with belief vectors from a dataset of visited POIs, then tasked with predicting beliefs for a disjoint set of POIs. For each test POI, we compute the mean absolute error (MAE) between the agent's predicted belief and the ground-truth value, reporting results separately across five semantic POI categories. Figure~\ref{fig:belief} presents category-wise belief estimation error for all evaluated models. As shown in Figure~\ref{fig:belief}, larger models like GPT-4o achieve the lowest MAE across all categories, followed by GPT-4o mini and Qwen-14B. ToolLLaMA excels in \emph{Transport} and \emph{Shops}, while smaller LLaMA-2 models show higher errors, especially for \emph{Entertainment}. Overall, larger LLMs generalize beliefs more accurately in diverse urban contexts.

\subsection{Ablation Study}

We conduct a systematic ablation study to assess the specific contribution of each architectural module to agent performance. As shown in Table~\ref{tab:ablation_citysim}, we remove key components --- belief module, recursive daily planning, goal module, needs module, and persona module, from our full framework and evaluate their effect on human-likeness scores across all domains.

Removing the \textbf{belief module} leads to a marked drop in performance, especially for activities and event reactions. Agents without beliefs lack the ability to accumulate or leverage prior expectations about places, resulting in less adaptive and less contextually grounded plans. This experiment demonstrates the critical role of beliefs in enabling value-driven, experience-aware decisions.

Ablating \textbf{recursive planning} reduces scores further, particularly in activity and mobility. Without this mechanism, agents are less capable of adapting routines to new internal or environmental feedback, which in turn diminishes the coherence and flexibility of their schedules.

We observe that excluding the \textbf{long-term goal module} has a more localized impact, with the most noticeable decline in dialogue and mobility. This suggests that long-term, high-level goal management primarily benefits life-course consistency and contextual coherence over multiple days.

Disabling the \textbf{needs module} results in the largest drop in both activity and event reaction scores. Agents without explicit needs prioritization become unable to interrupt or reorder their plans in response to internal states, often leading to unrealistic, rigid behavior and missed opportunities to fulfill basic requirements.

Finally, removing the \textbf{persona module}—thereby eliminating demographic and psychological diversity—causes a sharp decrease in all metrics. Agents converge to a bland, homogenized pattern, lacking the individualized routines and reactions essential for human-likeness.

\end{document}